\documentclass[sigconf]{acmart}

\usepackage{mdframed}
\usepackage{caption}
\usepackage{ragged2e} 
\usepackage{tabularx}
\usepackage{tcolorbox}
\usepackage{enumitem}
\usepackage{booktabs}
\usepackage{makecell}
\usepackage{amsmath}

\definecolor{myred}{RGB}{220, 50, 47}
\definecolor{mypurple}{RGB}{108, 113, 196}

\newcommand{\extInconsistency}[1]{\textcolor{myred}{#1}}
\newcommand{\intInconsistency}[1]{\textcolor{mypurple}{#1}}

\AtBeginDocument{%
  }

\acmConference[SCAI]{Proceedings of the SIGIR Workshop on Search-Oriented Conversational AI}{2026}{Melbourne, Australia}
\begin{document}

\title{Towards Detecting Inconsistencies in End-to-end Generated TODs}

\author{Tiziano Labruna}
\email{tlabruna@fbk.eu}
\orcid{0000-0001-7713-7679}
\affiliation{%
  \institution{Fondazione Bruno Kessler}
  \city{Povo, Trento}
  \country{Italy}
}

\author{Giovanni Bonetta}
\email{gbonetta@fbk.eu}
\orcid{0000-0003-4498-1026}
\affiliation{%
  \institution{Fondazione Bruno Kessler}
  \city{Povo, Trento}
  \country{Italy}
}

\author{Bernardo Magnini}
\email{magnini@fbk.eu}
\orcid{0000-0002-0740-5778}
\affiliation{%
  \institution{Fondazione Bruno Kessler}
  \city{Povo, Trento}
  \country{Italy}
}


\renewcommand{\shortauthors}{Labruna et al.}

\begin{abstract}
Generative AI is profoundly transforming the core technologies behind conversational systems, shifting from component-based to end-to-end approaches. However, Large Language Models (LLMs) may still generate inconsistencies, a critical issue particularly in Task-Oriented Dialogues (TODs), where system responses must strictly adhere to information from a domain knowledge base (e.g., restaurants in a city). A single hallucination (e.g., suggesting a non-existent restaurant) can lead to severe task failures.
We investigate a method for automatically detecting inconsistencies by conceptualizing TODs as a Constraint Satisfaction Problem (CSP), where variables represent dialogue segments referencing the conversational domain, and constraints among variables capture dialogue properties such as turn coherence and adherence to domain knowledge.
We propose a pipeline that first identifies variables in a target dialogue and then applies a CSP solver to identify valid solutions. By comparing the target dialogue with  valid variable assignments, we can detect inconsistencies and suggest minimal changes to ensure dialogue consistency.  We demonstrate the high accuracy of the CSP-based approach in detecting inconsistencies, and provide a detailed analysis of our findings.
\end{abstract}



\maketitle

\section{Introduction}
\label{introduction}

\begin{figure}[t]
    \centering

    \textbf{Knowledge Base} \\[0.3em]

    \setlength{\tabcolsep}{3pt} 
    \begin{tabularx}{\linewidth}{|c|X|X|X|X|}
        \hline
        \textbf{ID} & \textbf{Name} & \textbf{Area} & \textbf{Food} & \textbf{Price} \\
        \hline
        R1 & Taberna & centre & spanish & cheap \\
        R2 & Espana & centre & spanish & moderate \\
        R3 & Beirut & centre & lebanese & cheap \\
        \hline
    \end{tabularx}

    \vspace{1em}

    \textbf{Dialogue} \\[0.5em]
    \begin{mdframed}[backgroundcolor=gray!10, linewidth=0.5pt, innerleftmargin=8pt, innerrightmargin=8pt, innertopmargin=6pt, innerbottommargin=6pt]
    \begin{tabularx}{\linewidth}{@{}>{\bfseries}p{1.8cm}@{\hspace{0.5em}}X@{}}
        User: & I am looking for a restaurant serving \textbf{Spanish} food. \\
        System: & There are \intInconsistency{three} restaurants serving \textbf{Spanish} food, one is \textbf{cheap} and the other is \textbf{moderate} price range. Which price range would you prefer? \\
        User: & I am looking for a \textbf{cheap} restaurant in \textbf{any} area that serves \textbf{Spanish} food. \\
        System: & \extInconsistency{\textbf{Beirut}} is \textbf{cheap} and serves \extInconsistency{Lebanese} food. Would you like the location information? \\
    \end{tabularx}
    \end{mdframed}

    \caption{An inconsistent task-oriented dialogue (TOD) and a restaurant KB with slot-value pairs (N = Name, A = Area, F = Food, P = Price). Bold text highlights slot values in the dialogue. \extInconsistency{Red} values indicate dialogic inconsistencies, while \intInconsistency{purple} values indicate inconsistencies with the KB.}
    \label{fig:diag_example}
\end{figure}

Task-oriented dialogue (TOD) systems \cite{Balaraman-Survey-21,BUDZIANOWSKI-multiWOZ2018,mctear2020conversational,qin2023end} play a crucial role in human-computer interaction, facilitating seamless communication between users and machines to perform specific tasks. In recent years, transformer-based neural models have become the core technology behind TODs. In particular, pre-trained large language models (LLMs) allow end-to-end approaches \cite{bang2023multitask,lai2023external,qin2023end} that greatly simplify the development of conversational systems, with respect to more complex component-based pipelines \cite{young2013pomdp}. However, despite their impressive generative capabilities, it is well known that LLMs exhibit significant limitations in producing outputs that adhere to the requirements of task-specific domains \cite{cho-etal-2022-know,ji2022survey}. 
In a recent study \cite{labruna2024you} it has been shown that, when asked to generate a dialogue according to a given knowledge base ($KB$), as required by TODs, state-of-the-art open source LLMs produce up to 59\% of per dialogue disalignments with respect to the underlying KB.
Failing to align their outputs with a domain $KB$, leads to inconsistencies that undermine LLMs reliability in real-world applications.

Figure \ref{fig:diag_example} shows an example of a fragment of a Knowledge Base (three restaurants in a city) and a short TOD dialogue generated by a LLM. There are two hallucinations in this dialogue: first, at turn S1, the system mentions three restaurants serving Spanish food, which is not consistent with the knowledge base, where there are two such restaurants (this is a \textit{domain inconsistency}). Second, at turn S2, the system introduces a Lebanese restaurant, which, although existing in the $KB$, it is not coherent with the previous dialogue turns, as a Spanish restaurant would have been expected (this is a \textit{dialogic inconsistency}). Intuitively, both  domain and dialogic inconsistencies need the whole dialogue context in order to be detected: for instance, \textit{Lebanese} appears inconsistent because the user is looking for a Spanish restaurant since the beginning of the conversation, while considering turn S2 alone would result in a well formed dialogue.
In addition, notice that three changes would make the whole dialogue consistent: (i) changing \textit{three} with \textit{two} at turn S1; (ii) changing \textit{Lebanese} with \textit{Spanish} at turn S2; and (iii) changing \textit{Beirut} with \textit{Taberna} at turn S2. Detecting TOD inconsistencies and, if possible, suggesting how to solve them, is the goal of this paper.
The novel intuition of the paper is to consider dialogue consistency as a kind of \textit{Constraint Satisfaction Problem} (\textit{CSP} \cite{BRAILSFORD1999557}), under the following working hypothesis: (i) first, dialogue consistency can be modeled with a limited number of domain independent constraints that need to be respected by appropriate linguistic realizations; (ii) such constraints can be well represented to define a CSP, whose allowed solutions can be identified by a CSP solver; (iii) a TOD is consistent if its linguistic realizations belong to the set of solutions allowed by a CSP solver for that dialogue. In the paper, we discuss how dialogue constraints are defined, how they can be extracted and modeled as a CSP, and how to set up an experimental setting where we can empirically prove that a CSP solver can detect inconsistencies in a dialogue and suggest possible changes that make the dialogue consistent.

The contributions of the paper are the following:
\begin{itemize}
    \item We model TOD consistency as a Constraint Satisfaction Problem (CSP): to the best of our knowledge, this is a fully original approach.
    \item We set up a reusable experimental setting where TOD consistency can be automatically evaluated against a CSP solver.
    \item We show that the proposed CSP approach allows for effective detection of inconsistent TODs, achieving an accuracy of 75.9\%.
\end{itemize}


\section{Dialogue Consistency as a Constraint Satisfaction Problem}

In this section, we explore the conceptualization of dialogue consistency in the CSP framework. We first describe the fundamental component of a conversational domain (Section \ref{conversational_domain}), then we elucidate the various constraints that contribute to dialogue coherence (Section \ref{tod_consistency}), encompassing linguistic, dialogic, and domain-based considerations. We finally expound upon the formalization of dialogue constraints as CSPs (Section \ref{csp}), delineating the process of modeling dialogue coherence as a constraint satisfaction task.

\subsection{TOD Conversational Domain}
\label{conversational_domain}

TODs typically need specific knowledge about the conversational domain (e.g., a database of restaurants, a playlist of songs, etc.). As in literature \cite{henderson-etal-2014-second}, we assume a domain ontology providing a schema of the concepts (e.g., \textsc{Restaurant}, \textsc{Hotel}, \textsc{Movie}), a  set of slots $S$ (e.g., \textsc{Food}, \textsc{Area}, \textsc{Price}) for the concepts, and the set of values that each slot can assume (e.g., \textsc{Expensive}, \textsc{Moderate}, and \textsc{Cheap} for the \textsc{Price} slot). 
Then, a domain knowledge base ($KB$) comprises a collection of instances for the ontology concepts,  each  consisting of $[slot-value]$ pairs, adhering to the domain ontology schema.

\begin{figure*}
  \centering \includegraphics[height=0.17\textwidth]{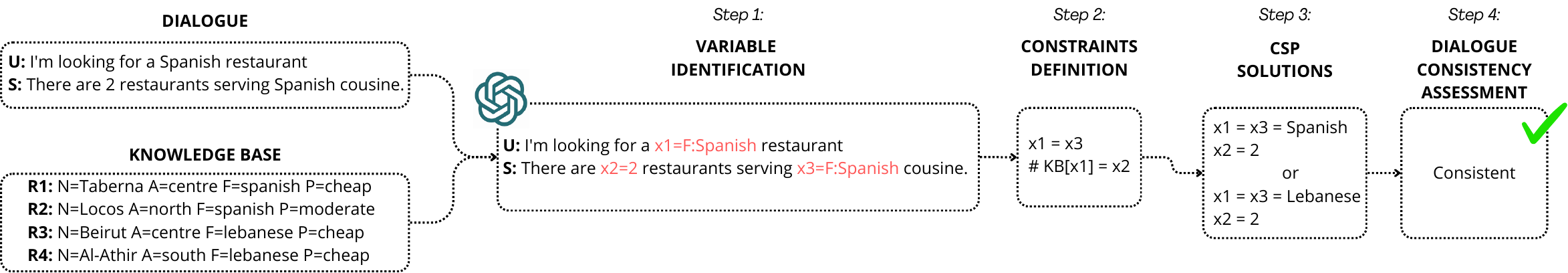}
  \caption{Overview of the CSP-based  methodology. In step 1 GPT-4o is used to annotate the given dialogue for variable identification. Step 2 allocates the constraints that need to be true for the dialogue to be consistent. Step 3 uses CSP to find the possible solutions, and in step 4 the original dialogue is matched with the CSP solutions to assess its correctness.}
  \label{fig:process}
\end{figure*}

\subsection{Dialogue Consistency}
\label{tod_consistency}
A TOD can be considered as a sequence of conversational turns between a user and a system aimed at achieving a specific goal. Within this framework, ensuring the consistency of the dialogue is crucial for effective communication between the user and the system. We consider three types of constraints, which need to be respected for a dialogue to be consistent: linguistic, dialogic and domain constraints. Figure~\ref{fig:restaurant-dialogue} provides a concrete example of how violations of these constraints can lead to inconsistencies, showing a case where the system response contradicts the underlying knowledge base despite a seemingly coherent dialogue flow.

\begin{figure}[htbp]
\centering
\begin{tabularx}{0.95\linewidth}{|c|X|X|X|X|}
    \hline
    \textbf{ID} & \textbf{Name} & \textbf{Area} & \textbf{Food} & \textbf{Price} \\
    \hline
    R1 & Mario & east & italian & expens. \\
    R2 & Napoli & centre & italian & cheap \\
    \hline
\end{tabularx}

\vspace{1em}

\begin{tcolorbox}[colback=gray!10, colframe=black!50, boxrule=0.5pt, left=8pt, right=8pt, top=6pt, bottom=6pt]
    \setlength{\itemsep}{0pt}
    \setlength{\parskip}{0pt}
    \begin{description}[leftmargin=1.5cm, labelwidth=1cm, labelsep=0.2cm]
        \item[U] I’m looking for a restaurant serving \textbf{\texttt{<MASK>}} food in any area.
        \item[S] There are no \textbf{\texttt{<MASK>}} restaurants in the area.
        \item[U] Well, can I get the phone number to a \textbf{\texttt{<MASK>}} restaurant?
        \item[S] Restaurant R2 serves \textbf{\texttt{<MASK>}} food. Their phone number is 01223 355166. Can I help you with anything else?
        \item[U] That’s it, goodbye.
        \item[S] You’re welcome, goodbye.
    \end{description}
\end{tcolorbox}

\vspace{1em}

\begin{tcolorbox}[colback=gray!5, colframe=black!40, boxrule=0.5pt, left=8pt, right=8pt, top=6pt, bottom=6pt]
    \setlength{\itemsep}{0pt}
    \setlength{\parskip}{0pt}
    \begin{description}[leftmargin=1.5cm, labelwidth=1cm, labelsep=0.2cm]
        \item[U] I’m looking for a restaurant serving \textbf{italian} food in any area.
        \item[S] There are no \textbf{italian} restaurants in the area.
        \item[U] Well, can I get the phone number to a \textbf{cheap} restaurant?
        \item[S] Restaurant R2 serves \textbf{italian} food. Their phone number is 01223 355166. Can I help you with anything else?
        \item[U] That’s it, goodbye.
        \item[S] You’re welcome, goodbye.
    \end{description}
\end{tcolorbox}

\caption{Example of an inconsistent re-lexicalization. The system is claiming there are no Italian restaurant, even though there are 2 in the Knowledge Base.} 
\label{fig:restaurant-dialogue}
\end{figure}

\paragraph{Linguistic Constraints.} They are necessary to respect general rules of language, including morpho-syntactic rules (e.g., genre and number agreement) and syntax-based rules (e.g., the correct use of a preposition). For instance, consider the following masked utterance:
\begin{tcolorbox}[colback=gray!10, colframe=black!50, boxrule=0.5pt, left=8pt, right=8pt, top=6pt, bottom=6pt]
    \begin{description}[leftmargin=2.5cm, labelwidth=1cm, labelsep=0.2cm]
        \item[U] I look for a restaurant in \textbf{\texttt{<MASK>}}.
    \end{description}
\end{tcolorbox}

\noindent The choice of \textit{centre} as a substitute for the masked token is valid, whereas \textit{expensive} would not be suitable because the preposition \textit{in} is rarely used to introduce a price in English.

\paragraph{Dialogic Constraints.} They  maintain the semantic coherence across successive turns of the dialogue, ensuring that each utterance logically aligns  with the preceding context, thereby facilitating a seamless flow of information. As an example, suppose the following masked dialogue turns:

\begin{tcolorbox}[colback=gray!10, colframe=black!50, boxrule=0.5pt, left=8pt, right=8pt, top=6pt, bottom=6pt]
    \begin{description}[leftmargin=2.5cm, labelwidth=1cm, labelsep=0.2cm]
        \item[U] I would like an Italian restaurant.
        \item[S] There is no \textbf{\texttt{<MASK>}} restaurant in the centre.
    \end{description}
\end{tcolorbox}

\noindent
Here both \textit{Italian} and \textit{cheap} would be eligible choices from a linguistic point of view, but only \textit{Italian} would maintain the coherence with the previous turn in the dialogue.

\paragraph{Domain Constraints.} They  ensure alignment between the dialogue content and the  knowledge base of the system, thereby maintaining the dialogue's alignment with relevant factual information. Consider, for instance, a $KB$ with the following restaurants:

\vspace{0.5em}

\begin{tabularx}{0.95\linewidth}{|c|X|X|X|X|}
    \hline
    \textbf{ID} & \textbf{Name} & \textbf{Area} & \textbf{Food} & \textbf{Price} \\
    \hline
    R1 & Mario & east & italian & expens. \\
    R2 & Napoli & centre & italian & cheap \\
    \hline
\end{tabularx}

\vspace{0.5em}
\par
\noindent
And the following piece of masked dialogue:

\begin{tcolorbox}[colback=gray!10, colframe=black!50, boxrule=0.5pt, left=8pt, right=8pt, top=6pt, bottom=6pt]
    \begin{description}[leftmargin=2.5cm, labelwidth=1cm, labelsep=0.2cm]
        \item[U] I am looking for an \textbf{Italian} restaurant in the \textbf{centre}.
        \item[S] We have \textbf{\texttt{<MASK>}} restaurants available for your preferences.
    \end{description}
\end{tcolorbox}

Then, the only admissible choice for the masked token would be \textit{one}, as selecting any other number would introduce an inconsistency with the information provided in the $KB$.

\subsection{TOD Consistency as CSP}
\label{csp}

A Constraint Satisfaction Problem (CSP) \cite{BRAILSFORD1999557,kumar1992algorithms} defines a set of variables, each associated with a finite domain of values, and a set of constraints specifying the allowed combinations of values. A solution is an assignment of values to all variables that satisfies all constraints; if no such assignment exists, the CSP is unsolvable.

We model TOD consistency as a CSP, where variables correspond to dialogue elements to be instantiated (e.g., $MASK$ tokens), and their domains are derived from the dialogue-specific $KB$. Constraints capture linguistic, dialogic, and domain requirements (Section \ref{tod_consistency}). 

Formally, given a dialogue $d_i$ with variables $x_1, \ldots, x_n$, domains $D_1, \ldots, D_n$, and a set of constraints $\mathcal{C}$, the task is to determine whether there exists an assignment $A = \{(x_1, a_1), \ldots, (x_n, a_n)\}$ with $a_i \in D_i$ such that:
\[
\text{\textit{Satisfies}}(A, C_j) \quad \forall C_j \in \mathcal{C}
\]
where $\text{\textit{Satisfies}}(A, C_j)$ indicates whether assignment $A$ satisfies constraint $C_j$.

\section{Methodology}
\label{methodology}
This section outlines the process of modeling a TOD as a CSP, and then to assess the dialogue consistency using a CSP solver. The assessment involves three key steps for a   $[d, KB]$ pair, where $d$ is a dialogue  and $KB$ is a Knowledge Base: (1) identification of the variables within the dialogue $d$ (Section \ref{id_var}); (2) definition of dialogue constraints and construction of a CSP solver for the  $[d, KB]$ pair (Section \ref{constraints}); and (3) application of the CSP solver to determine if the dialogue $d$ represents a feasible solution with respect to the defined constraints (Section \ref{assess}). These phases of the methodology are illustrated in Figure \ref{fig:process}.

\subsection{Identifying TOD Variables}
\label{id_var}

At step 1 (see Figure \ref{fig:process}), we consider a TOD $d$ and a $KB$ (i.e., a set of entities described by slot-value pairs) related to the conversational domain of the dialogue. We do not assume any particular dependency between $d$ and $KB$: $d$ could be either fully covered by $KB$ (i.e., all mentions of slot values in $d$ are present in $KB$), only partially covered, or not covered at all. 
We consider CSP variables all text portions in $d$ either referring  to a slot value in $KB$ or mentioning amounts of instances in $KB$. The rationale is that both slot values and instance amounts are elements that better characterize a TOD and are responsible for its consistency. In our example in Figure \ref{fig:diag_example}, we will obtain the following variables with their assignments, corresponding to highlighted tokens:

$[x_1 = Spanish]$, $[x_2 = three]$, $[x_3 = Spanish]$, $[x_4 = cheap]$ ... $[x_{11} = Lebanese]$.

\subsection{Defining TOD Constraints}
\label{constraints}

We have  established a set $\mathcal{X}$ of variables $x_1, x_2, ..., x_n$, where each variable $x_i$ can assume a value either from the slot values or from amounts of instances  in $KB$. Moving to step 2  in Figure \ref{fig:process}, we now define the set of constraints $\mathcal{C}$ over the values that can be assigned to $\mathcal{X}$ variables. 
We consider the three categories of constraints introduced in Section \ref{tod_consistency}: linguistic, dialogic, and domain-based constraints, and for each category we define a set of domain independent patterns, which are then instantiated as actual constraints on a  TOD.

\paragraph{Patterns for linguistic constraints.}
 We model linguistic constraints as the need for a variable derived from a slot value to match the semantic type of its slot type.
 For instance, given the utterance \textit{I am looking for a restaurant at \textsc{$x_1$}}, the  value of the variable \textsc{$x_1$} must belong to the \textsc{Area} type. More precisely, $C1$ is defined as follows:
\[C1: x_1 \in V\]
where $V$ is the set of values belonging to the same slot type as the original value.
Constraint $C1$, is meant to avoid that a variable can assume values that are semantically non valid. For instance, avoiding that \textsc{$x_1$=north} can be assigned to a \textsc{Food}, as in \textit{I am looking for a restaurant at \textsc{indian}}, which is ungrammatical in English.

\paragraph{Patterns for dialogic constraints.}
We consider two dialogic constraints. $C2$ ensures that variables referring to the same slot-name and slot-value in $d$ are assigned to the same value. $C3$ ensures that variables with the same semantic type (i.e., same slot-name) occurring in the same utterance are assigned to different values. 
Given the turn \textit{U: I want an $x_1$ restaurant. S: There are 3 restaurant that serve $x_2$}, we define $C2$ as follows:
\[C_2: x_1 = x_2\]

\noindent
where the aim is to keep internal coherence across the dialogue turns. 
Given the utterance \textit{We have $x_1$, $x_2$, or $x_3$ restaurants.}, we define $C3$ as:
\[
C_3: x_1 \neq x_2, \quad x_1 \neq x_3, \quad x_2 \neq x_3
\]

\noindent
which captures non redundancy at the utterance level.

\paragraph{Patterns for domain-based constraints.} We consider three domain-based constraints. All of them are meant to guarantee consistency between the number of instances mentioned in $d$ and the actual number of instances present in $KB$. We distinguish three cases: $C4$ covers the cases when an utterance in $d$ states that there are no instances in $KB$; $C5$ covers the cases where it is stated that there is at least one instance; and $C6$ the cases where there are exactly $n$ instances.

As for $C4$, consider an utterance indicating no results for a search: \textit{There are no restaurants serving $x_1$ food}, assuming that there are no restaurants with \textsc{[Food=$x_1$]} in $KB$. For this utterance, $C4$ is defined as:
\[C4: \neg \exists i \in KB \text{ with values } x_1\]
implying that the variable $x_1$ can not assume a value that is present in an instance of the $KB$.

As for $C5$, consider the utterance:
 \textit{We have many $x_1$ restaurants at $x_2$}, where at least one restaurant with \textsc{[Food=$x_1$]} and \textsc{[Area=$x_2$]} is supposed to exist in $KB$. For this utterance, $C5$ is defined as:
\[C5: \exists i \in KB \text{ with values } x_1, x_2\]
imposing the existence of at least one instance with values $x_1$ and $x_2$.
Finally, for $C6$, consider the utterance \textit{There are $x_1$ restaurants at $x_2$}. We define the constraint as:
\[C6: |\{i \in KB \text{ with value }x_2\}| = x_1\]
to check that the number of instances with value $x_2$ is exactly equal to $x_1$. 

To sum up, we have defined six general, domain independent (i.e., in principle they can be applied to any TOD), constraint patterns over the variable of a TOD. 

\subsection{Assessing Dialogue Consistency}  
\label{assess}  

After identifying all variables and constraints for a dialogue \(d\), a CSP solver computes all possible solutions for the variables in \(d\) based on the knowledge base (\(KB\)) (step 3 in Figure \ref{fig:process}). If one of these solutions matches the variable assignments in \(d\), the dialogue is consistent with \(KB\) (step 4 in Figure \ref{fig:process}).  For example, in Figure \ref{fig:diag_example}, the assignment \([x_2 = \text{three}]\) violates \(C6\) (incorrect count of Spanish instances in \(KB\)), while \([x_{11} = \text{Lebanese}]\) violates \(C2\) (lack of coherence with prior turns). If the CSP solver finds at least one solution, the variable assignments in the dialogue must match one of those solutions to ensure all constraints are satisfied. Conversely, if no solution exists with respect to \(KB\), the variable assignments should either remain empty or include values not present in \(KB\) to maintain consistency. When at least one solution is found but none matches the variable assignments, the solver identifies the most similar solution and determines the minimal changes required to make the dialogue consistent. This process provides a detailed report highlighting specific inconsistencies and suggesting corrections.

\section{Validating CSP Performance}
\label{sec:assessing-performance}

We evaluate the ability of the CSP-based approach to detect inconsistencies in task-oriented dialogues (TODs) through a controlled experiment (Figure \ref{fig:process}). We construct a balanced dataset of dialogue--knowledge base pairs \([d, KB]\), with equal proportions of \textsc{consistent} and \textsc{not-consistent} instances. Each pair is processed through the pipeline, and the CSP component outputs a binary decision, which is compared against ground truth.

\subsection{Experimental Setup}
\label{sec:exp_setup}

We construct a balanced dataset of 108 dialogue--KB pairs from MultiWOZ 2.3. While the dataset is assumed to be consistent, manual inspection combined with CSP verification revealed that approximately 10\% of dialogues exhibit inconsistencies due to annotation errors; these were removed.
The remaining dialogues were split evenly into \textsc{consistent} and \textsc{not-consistent} subsets. Inconsistent dialogues were generated by randomly modifying slot values to violate KB constraints. For each dialogue, a tailored $KB$ was constructed by selecting relevant entities from the global MultiWOZ knowledge base.

CSP variables are identified either from MultiWOZ annotations or via GPT-4o, and constraints are instantiated using the six patterns introduced in Section \ref{constraints}. 
We model the problem using MiniZinc \cite{nethercote2007minizinc}, a declarative constraint programming language. 
We use the Chuffed solver \cite{chu2018chuffed}, which is optimized for constraint satisfaction problems and supports efficient search and propagation strategies. The solver determines whether a valid assignment exists; absence of a solution implies inconsistency.

We compare four classification methods: a random baseline, two CSP-based approaches using MultiWOZ annotations (global and local), and a fully automated pipeline using GPT-4o for variable extraction. Performance is measured using accuracy over \([d, KB]\) pairs. 

We consider four methods for classifying dialogue consistency:

\begin{itemize}
    \item \textsc{Random Baseline}: assigns labels randomly (expected accuracy: 50\%).
    \item \textsc{MWoZ Global + CSP}: variables are extracted from full-dialogue annotations and evaluated globally.
    \item \textsc{MWoZ Local + CSP}: variables are evaluated independently per turn; a single inconsistent turn marks the dialogue as inconsistent.
    \item \textsc{GPT-4o Global + CSP}: variables are extracted automatically using a two-step prompt chain, then evaluated globally.
\end{itemize}

The prompt chain used to annotate the dialogue turns consists of the following two prompts:

\begin{itemize}
    \item \textbf{Prompt-1:} Analyze the given user utterance and extract any slot-value pairs. The possible slot types are: Area, Food, Price, Depart, Destination. Return the output as JSON with the dialog-act format.
    \item \textbf{Prompt-2:} Refine the given annotation for the user utterance. Ensure that only slots related to Area, Food, and Price are included. Correct any errors in the provided annotation, add missing slots, and remove any irrelevant slots. Return the output as JSON with the updated dialog-act format.
\end{itemize}

The GPT-4o response to Prompt-1 is used as input within Prompt-2, and the final output is a JSON file containing annotations about slot variables.

The \textit{dialog-act} referred to in the two prompts is a JSON schema that guides GPT-4o in structured output mode and resembles the MultiWOZ JSON annotation schema.

\begin{table}[ht]
\centering
{\small
\setlength{\tabcolsep}{4pt} 
\begin{tabular}{l c}
\toprule
\textbf{Method} & \textbf{accuracy (\%)} \\
\midrule
\textsc{Random Baseline} & 50.0 \\
\midrule
\textsc{MWoZ global variables + CSP} & 91.6 \\
\textsc{MWoZ local variables + CSP} & 79.0 \\
\textsc{GPT4-o global variables + CSP} & 75.9 \\
\bottomrule
\end{tabular}
}
\caption{Results on assessing CSP performance.}
\label{tab:performance}
\end{table}

\subsection{Results}
\label{results-1}

Table \ref{tab:performance} reports the results. The upper-bound method based on MultiWOZ global annotations achieves 91.6\% accuracy, confirming the effectiveness of the CSP formulation and the high coverage of the constraint patterns. Performance drops to 79\% when constraints are applied locally, highlighting the importance of global context. The end-to-end pipeline using GPT-4o reaches 75.9\%, with errors primarily due to imperfect variable extraction.
The results show that (i) CSP-based modeling is effective for detecting dialogue inconsistencies, and (ii) the proposed constraint set provides strong coverage of relevant phenomena.

\section{Analysing LLM Behavior}
\label{sec:llm_behavior}

We investigate how large language models (LLMs) handle dialogue consistency under explicit constraints. Specifically, we study (i) their ability to generate consistent TODs, (ii) the role of different constraint types, and (iii) the localization of inconsistency sources.

\subsection{Experimental Setup}
\label{sec:llm_setup}

We consider 950 MultiWOZ dialogues across multiple domains. Each dialogue is de-lexicalized by replacing slot values with placeholders. Models are then prompted to reconstruct the original dialogue by filling these placeholders using the provided Knowledge Base ($KB$), thereby generating a re-lexicalized dialogue.

All models operate in a zero-shot setting without any fine-tuning. Closed-source models are accessed via APIs, while open-source models are run using HuggingFace checkpoints.

The full prompt used in our experiments is reported below.

\textbf{System Prompt:}
\begin{quote}
You are given an instruction that outlines a task, a Knowledge Base containing domain-specific information, and a dialogue to process.

Your goal is to fill in the \texttt{[MASK]} placeholders in the dialogue using only the information provided in the Knowledge Base.

\textbf{Task:}
Replace each \texttt{[MASK]} with the most appropriate value from the Knowledge Base. Preserve the original structure of the dialogue exactly. If a turn does not contain any placeholders, leave it unchanged. Each dialogue turn must start with either \textit{User} or \textit{System}. Maintain the original spacing and punctuation (e.g., write ``Hi ,'' instead of ``Hi,'').
\end{quote}

\textbf{Input Format:}
\begin{quote}
\begin{verbatim}
[Knowledge Base]

<key-value pairs>

[Dialogue]

User: ...
System: ...
User: ...
...
\end{verbatim}
\end{quote}
\textbf{Output:}
\begin{quote}
The same dialogue with all [MASK] tokens replaced accordingly.

\end{quote}

The generated dialogue is then evaluated using the CSP solver to verify constraint satisfaction.

\subsubsection{Models}
We evaluate four language models: LLaMA-3.1 8B, GPT-3.5-Turbo, GPT-4o, and GPT-o1. LLaMA-3.1 8B is a large-scale model fine-tuned for handling complex dialogue contexts and maintaining coherence in text generation \citep{dubey2024llama}. GPT-3.5-Turbo is a model specifically designed for conversational tasks \citep{achiam2023gpt}. GPT-4o is an advanced language model recognized for its robust performance in various natural language processing tasks \citep{hurst2024gpt}. GPT-o1 is one of the latest update of the GPT series, designed to  reason through complex tasks to solve harder problems\footnote{https://openai.com/o1/}. All models were prompted with both the de-lexicalized dialogue, $d_{delex}$, and its associated $KB$ as input, ensuring a comprehensive context for producing dialogues that adhered to implicit constraints. Inference was conducted in zero-shot mode without fine-tuning, leveraging the respective APIs for closed source models and the huggingface checkpoints for the open ones: GPT-3.5-Turbo (2023-05-15), GPT-4o and GPT-o1 (2024-05-13), and LLaMA-3.1 8B (2023-07-10).

\subsubsection{Baselines}
To comparison, we included four non-trivial dialogue re-lexicalization baselines:
\begin{itemize}
    \item \textsc{Random-ALL} generates a re-lexicalized dialogue $d_{relex}$ by randomly assigning variables in $d_{delex}$ to any slot values present in the $KB$, regardless of their slot type.
    \item \textsc{Random-SLOT} also assigns variables randomly but restricts the selection to values associated with the same slot type as the original.
    \item \textsc{Most Frequent-ALL} baseline assigns variables in $d_{delex}$ to the most frequent slot values found across all slots in the $KB$.
    \item \textsc{Most Frequent-SLOT} baseline selects the most frequent value from the same slot type as the original.
\end{itemize}

\subsubsection{Evaluation Metrics}
We use \textit{Global Consistency Accuracy} (GCA) and \textit{Variable Consistency Accuracy} (VCA) as the metrics to evaluate the adherence of a dialogue to a specific set of constraints. Given a re-lexicalized dialogue $d_{relex}$ where CSP variables are assigned to values, GCA measures the overall accuracy of the assignments for each variable. The average GCA is calculated as the proportion of dialogues that fully comply with all defined constraints:
\[ GCA = \frac{\sum_{i=1}^{N} \left( \prod_{j=1}^{M} \textit{Satisfies}(A_i, C_j) \right)}{N} \]
where \(N\) is the total number of dialogues, and \(\textit{Satisfies}(A_i,C_j)\) is a binary indicator function that returns 1 if and only if all variable assignments in dialogue \(d_i\) comply with the constraint $j$, 0 otherwise.
On the other hand, VCA assesses the assignment accuracy on individual variables within the dialogue. We compare the dialogue assignment to the solutions of the CSP solver and find the most similar solution; then, we count how many variable assignments coincide with the assignments of the most similar solution. We formally define VCA as follows:
\[ VCA = \frac{\sum_{i=1}^{N} \lvert \textit{CorrectAssignments}(d_i) \rvert}{M} \]

\noindent
where \(N\) is the total number of dialogues, \(M\) is the total number of variables in the dialogues, and \(\textit{CorrectAssignments}(d_i)\) are the variable assignments in dialogue \(d_i\) that coincide with the assignments of the most similar solution provided by the CSP solver.
GCA and VCA provide insights into the ability of the dialogue generation system to maintain coherence and fidelity to the underlying domain knowledge while generating responses. Higher values of GCA and VCA indicate better performance in terms of dialogue quality and consistency, unlike traditional dialogue evaluation metrics (e.g., BLEU, ROUGE, or perplexity).

Additionally, the process used for computing VCA can be extended to identify specific errors within a dialogue. In cases where a dialogue is not among the solutions identified by the CSP, the most similar solution can be used to detect erroneous slot-value assignments. Specifically, errors are defined as slot-values that, if corrected, would result in a solution satisfying all constraints. This enables the generation of detailed reports pinpointing the errors in the dialogue, facilitating more targeted improvements.

\begin{table}[ht]
\centering
\begin{tabular}{l c c}
\toprule
 \textbf{Method} & \textbf{GCA} & \textbf{VCA}\\
\midrule
\textsc{Random-ALL} & 0.01 & 0.02\\
\textsc{Random-SLOT} & 0.01 & 0.12\\
\textsc{Most Frequent-ALL} & 0.01 & 0.11\\
\textsc{Most Frequent-SLOT} & 0.06 & 0.23\\
\textsc{Llama-3.1 8B} & 0.03 & 0.08\\
\textsc{GPT-3.5-turbo} & 0.11 & 0.37\\
\textsc{GPT-4o} & \textbf{0.14} & 0.41\\
\textsc{GPT-o1} & \textbf{0.14} & \textbf{0.42}\\
\bottomrule
\end{tabular}
\caption{Baselines and model performance on re-lexicalizing TODs.}
\label{tab:baselines}
\end{table}

\subsection{Results}

Table \ref{tab:baselines} shows that GPT-4o and GPT-o1 outperform other models, although absolute performance remains moderate, indicating that constraint satisfaction is still challenging for LLMs. Performance improves as the number of valid CSP solutions increases, suggesting that dialogues with higher flexibility are easier to generate correctly.
The constraint analysis (Table \ref{tab:constraints}) highlights the importance of domain-level constraints, particularly \(C6\), in ensuring consistency. These findings indicate that while modern LLMs can partially capture structured constraints, explicit modeling via CSP remains beneficial for enforcing consistency and diagnosing errors.

\subsubsection{Ablation Study}
Table \ref{tab:ablation} presents the results of an ablation study we conducted.
The ablation study removes one constraint at a time to measure impact on GCA and VCA. Results indicate that \(C6\) (exact match with KB instances) is the most critical, followed by \(C1\) (hard constraints on slot values).

\begin{table}[h]
\centering
\begin{tabular}{l c c}
\toprule
\textbf{Constraint} & \textbf{GCA} & \textbf{VCA} \\
\midrule
\makecell[l]{\textsc{all except C1}} & 0.15 & 0.45 \\
\makecell[l]{\textsc{all except C2}} & 0.15 & 0.42 \\
\makecell[l]{\textsc{all except C3}} & 0.15 & 0.45 \\
\makecell[l]{\textsc{all except C4}} & 0.15 & 0.46 \\
\makecell[l]{\textsc{all except C5}} & 0.15 & 0.45 \\
\makecell[l]{\textsc{all except C6}} & \textbf{0.21} & \textbf{0.48} \\
\midrule
\makecell[l]{\textsc{all except} \\ \textsc{dialogic}} & 0.15 & 0.45 \\
\makecell[l]{\textsc{all except} \\ \textsc{domain}} & \textbf{0.23} & \textbf{0.56} \\
\bottomrule
\end{tabular}
\caption{Ablation study: global and variable consistency under different constraint configurations.}
\label{tab:ablation}
\end{table} 

\begin{table}[t]
\centering
\begin{tabular}{l c c}
\toprule
\textbf{Constraint} & \textbf{\# variables} & \textbf{\% coverage} \\
\toprule
\textsc{C1} & \textbf{9281} & \textbf{100\%}\\
\textsc{C2} & 6084 & 66\% \\
\textsc{C3} & 1124 & 12\% \\
\textsc{C4} & 301 & 3\% \\
\textsc{C5} & 2369 & 26\% \\
\textsc{C6} & 4257 & 46\% \\
\bottomrule
\end{tabular}
\caption{Number and proportion of variables affected by each constraint.}
\label{tab:constraints}
\end{table}

\subsubsection{Additional Results: CSP Solution Distribution}

\begin{table}[t]
\centering
\setlength{\tabcolsep}{4pt}
\begin{tabular}{l c c}
\toprule
\textbf{Dataset} & \textbf{\# dialogues} & \textbf{\# variables}\\
\midrule
\textsc{All} & 950 & 9281\\
\textsc{1 solution} & 18 & 54\\
\textsc{2-10 solutions} & 134 & 868\\
\textsc{11-100 solutions} & 286 & 2332\\
\textsc{101+ solutions} & 306 & 3151\\
\bottomrule
\end{tabular}

\caption{Dialogue distribution based on CSP solutions (MiniZinc).}
\label{tab:diag_sol}
\end{table}

Table \ref{tab:diag_sol} reports the distribution of dialogues based on the number of valid CSP solutions. Dialogues with a higher number of solutions tend to allow more flexibility in variable assignments, which correlates with improved LLM performance.

\section{Related Work}

TOD systems have been extensively investigated in NLP \cite{allen2001architecture}. 
Recent research has explored the use of neural network architectures for dialogue state tracking \cite{wu2020gcdst,zhao2021effective,labruna2023addressing} and policy learning \cite{su2016line,liu2017iterative}.
Several metrics have been proposed to assess the performance of TOD systems, including task completion rates, user satisfaction scores, and objective measures for system components, such as precision, recall, and F1-score \cite{chen2017survey,santhanam2019towards,deriu2021survey}. Recent studies have emphasized the importance of holistic evaluation frameworks that consider multiple aspects of dialogue quality \cite{zhang2021d,labruna2024you}.
Maintaining consistency and coherence in dialogues is essential for effective communication between users and dialogue systems. Previous research has investigated various approaches to ensure dialogue coherence, including coherence modeling ~\cite{cervone2018coherence}, and coherence-based response generation \cite{cervone2020dialogue}, aiming to enhance the naturalness and fluency of generated dialogues.
Finally, several studies have explored the application of CSPs to language. These include early attempts to ensure coherence in generated text \cite{kibble2004optimizing}, model preposition lexicalization using constraints \cite{moriceau2004constraint}, guide lexical choices through constraints \cite{mckeown1997floating}, and treat context-sensitive utterance generation as a CSP \cite{popescu2009constraint}. Differently to these works, our approach focuses on detecting inconsistencies in already generated TOD dialogues using CSP.


\section{Conclusion}
Generative LLMs may produce inconsistent TODs, due to misalignment between parametric memory and the TOD $KB$. We have introduced a novel approach to detect TOD inconsistencies based on Constraint Satisfaction. 
Several experiments  demonstrate the feasibility of the approach, enabling  to effectively identify and quantify inconsistencies present in  TODs with high accuracy (75.9\% with GPT-4o and CSP solver).
We also analysed the LLM inconsistencies when tasked to re-lexicalize TODs, finding that they primarily concern domain knowledge adherence, resulting in an overall accuracy of only 0.14 at the dialogue level. 
Our study highlights the potential of CSP-based methodologies in evaluating dialogue consistency and identifying areas for improvement in automated dialogue generation systems. Future research should further explore the application of CSP in task-oriented dialogues and investigate strategies to enhance the coherence of LLM-generated dialogues, particularly in applications with strong domain knowledge requirements.

\section*{Limitations}
\label{sec:limitations}

While the proposed Constraint Satisfaction Problem (CSP)-based approach offers a novel and effective method for detecting inconsistencies in task-oriented dialogues (TODs), it presents several limitations.

The system relies on the explicit mapping of dialogues into variable-constraint representations. Although our method is domain-independent in principle, the process of extracting variables and constraints from dialogues may require customization or adaptation for new domains or dialogue schemas.

Our method focuses on identifying inconsistencies and suggesting minimal changes for correction, but it does not automatically regenerate fluent or user-aligned responses after such modifications. This leaves the generation of corrected natural language utterances as future work.

Finally, although our experimental results are promising, they are based on controlled datasets and manually designed inconsistencies. Further work is needed to assess robustness in more complex or organically generated dialogues.

\section*{Ethical Considerations}
\label{sec:ethics}

\textbf{Use of Scientific Artifacts.}
We used publicly available task-oriented dialogue datasets for experimentation. These datasets include MultiWOZ \cite{BUDZIANOWSKI-multiWOZ2018} and variations based on it. Additionally, we used off-the-shelf large language models (LLMs) to generate new dialogues with intentional inconsistencies for controlled evaluation. All code developed for the CSP-based inconsistency detection pipeline is our original contribution and will be made publicly available for research purposes under an open-source license.

\textbf{Licensing and Intended Use.}
All external datasets and models used in this work were employed in accordance with their licenses. Our use was consistent with the intended purpose of the datasets (research), and we explicitly specify that the CSP-based system and associated data artifacts are intended solely for research and educational use. Any derivative dataset created using our framework also inherits this research-only restriction.

\textbf{Privacy and Data Integrity.}
The dialogue data used in this study does not include personally identifiable information (PII), and no effort was made to collect or infer such data. We manually verified the synthetic and benchmark dialogues for inappropriate or offensive content, and none was found. Our system does not involve any human annotation beyond the authors, so no consent or risk disclaimers were required.

\textbf{Documentation and Statistics.}
All artifacts, including the experimental codebase, constraint templates, and synthetic dialogue generation scripts, are provided with the submission. This includes coverage across domains, types of slot-value inconsistencies, and linguistic patterns. We report the number of dialogue examples used in each experiment, as well as their train/test splits, in the experimental section. We also provide accuracy scores as descriptive statistics for the evaluation.

\textbf{Computational Resources.}
Model generation and evaluation were conducted using a single NVIDIA A40 GPU, with a total budget of approximately 40 GPU hours. We do not fine-tune any large models; our work only uses them in inference mode.

\textbf{Use of Existing Software.}
Our system uses standard NLP libraries such as Hugging Face Transformers and MiniZinc `constraint` solver library. All packages were used with default or explicitly documented parameters.

\textbf{Human Participants.}
No human participants were recruited for this study, and no user studies or annotation tasks involving external contributors were conducted. Therefore, issues such as compensation or informed consent do not apply in our setting.

%
%

\bibliographystyle{ACM-Reference-Format}
\bibliography{sample-base}

@article{ji2022survey,
    title   = {Survey of hallucination in natural language generation},
    author  = {Ziwei Ji and others},
    journal = {ACM Computing Surveys},
    year    = {2022}
}

@inproceedings{henderson-etal-2014-second,
    title     = {The Second Dialog State Tracking Challenge},
    author    = {Matthew Henderson and others},
    booktitle = {Proceedings of the 15th Annual Meeting of the Special Interest Group on Discourse and Dialogue ({SIGDIAL})},
    month     = jun,
    year      = {2014},
    address   = {Philadelphia, PA, U.S.A.},
    publisher = {Association for Computational Linguistics},
    url       = {https://www.aclweb.org/anthology/W14-4337},
    doi       = {10.3115/v1/W14-4337},
    pages     = {263--272}
}

@article{BUDZIANOWSKI-multiWOZ2018,
  title = {MultiWOZ--A Large-Scale Multi-Domain Wizard-of-Oz Dataset for Task-Oriented Dialogue Modelling},
  author = {Budzianowski, Paweł and Wen, Tsung-Hsien and Tseng, Bo-Hsiang and Casanueva, Inigo and Ultes, Stefan and Ramadan, Osman and Gašić, Milica},
  journal = {arXiv preprint arXiv:1810.00278},
  year = {2018}
}

@article{mctear2020conversational,
  title={Conversational AI: Dialogue Systems, Conversational Agents, and Chatbots},
  author={McTear, Michael},
  journal={Synthesis Lectures on Human Language Technologies},
  volume={13},
  number={3},
  pages={1--251},
  year={2020},
  publisher={Morgan \& Claypool Publishers}
}

@article{young2013pomdp,
  title={Pomdp-based statistical spoken dialog systems: A review},
  author={Young, Steve and Ga{\v{s}}i{\'c}, Milica and Thomson, Blaise and Williams, Jason D},
  journal={Proceedings of the IEEE},
  volume={101},
  number={5},
  pages={1160--1179},
  year={2013},
  publisher={IEEE}
}

@inproceedings{Balaraman-Survey-21,
  author    = {Vevake Balaraman and
               Seyedmostafa Sheikhalishahi and
               Bernardo Magnini},
  title     = {Recent Neural Methods on Dialogue State Tracking for Task-Oriented
               Dialogue Systems: A Survey},
  booktitle = {Proceedings of the 22nd Annual Meeting of the Special Interest Group
               on Discourse and Dialogue, SIGdial 2021, Singapore and Online, July
               29-31, 2021},
  pages     = {239--251},
  publisher = {Association for Computational Linguistics},
  year      = {2021},
  url       = {https://aclanthology.org/2021.sigdial-1.25},
  bibsource = {dblp computer science bibliography, https://dblp.org}
}

@article{bang2023multitask,
  title={A multitask, multilingual, multimodal evaluation of chatgpt on reasoning, hallucination, and interactivity},
  author={Bang, Yejin and Cahyawijaya, Samuel and Lee, Nayeon and Dai, Wenliang and Su, Dan and Wilie, Bryan and Lovenia, Holy and Ji, Ziwei and Yu, Tiezheng and Chung, Willy and others},
  journal={arXiv preprint arXiv:2302.04023},
  year={2023}
}

@inproceedings{lai2023external,
  title={External knowledge acquisition for end-to-end document-oriented dialog systems},
  author={Lai, Tuan M and Castellucci, Giuseppe and Kuzi, Saar and Ji, Heng and Rokhlenko, Oleg},
  booktitle={Proceedings of the 17th Conference of the European Chapter of the Association for Computational Linguistics},
  pages={3633--3647},
  year={2023}
}

@inproceedings{labruna2023addressing,
  title={Addressing Domain Changes in Task-oriented Conversational Agents through Dialogue Adaptation},
  author={Labruna, Tiziano and Magnini, Bernardo},
  booktitle={Proceedings of the 17th Conference of the European Chapter of the Association for Computational Linguistics: Student Research Workshop},
  pages={149--158},
  year={2023}
}

@inproceedings{allen2001architecture,
  title={An architecture for more realistic conversational systems},
  author={Allen, James and Ferguson, George and Stent, Amanda},
  booktitle={Proceedings of the 6th international conference on Intelligent user interfaces},
  pages={1--8},
  year={2001}
}

@article{kumar1992algorithms,
  title={Algorithms for constraint-satisfaction problems: A survey},
  author={Kumar, Vipin},
  journal={AI magazine},
  volume={13},
  number={1},
  pages={32--32},
  year={1992}
}

@inproceedings{nethercote2007minizinc,
  author = {Nethercote, Nicholas and Stuckey, Peter J. and Becket, Rowan and Brand, Simon and Duck, Greg J. and Tack, Guido},
  title = {MiniZinc: Towards a standard CP modelling language},
  booktitle = {CP 2007},
  editor = {Bessiere, Christian},
  volume = {4741},
  series = {LNCS},
  pages = {529--543},
  publisher = {Springer},
  year = {2007},
  url = {http://www.minizinc.org/}
}

@misc{chu2018chuffed,
  author = {Chu, Geoffrey and Stuckey, Peter J. and Schutt, Anthony and Ehlers, Thorsten and Gange, Graeme and Francis, Keith},
  title = {Chuffed, a lazy clause generation solver},
  year = {2018},
  howpublished = {\url{https://github.com/chuffed/chuffed}}
}

@article{achiam2023gpt,
  title={Gpt-4 technical report},
  author={Achiam, Josh and Adler, Steven and Agarwal, Sandhini and Ahmad, Lama and Akkaya, Ilge and Aleman, Florencia Leoni and Almeida, Diogo and Altenschmidt, Janko and Altman, Sam and Anadkat, Shyamal and others},
  journal={arXiv preprint arXiv:2303.08774},
  year={2023}
}

@inproceedings{wu2020gcdst,
  title={GCDST: A graph-based and copy-augmented multi-domain dialogue state tracking},
  author={Wu, Peng and Zou, Bowei and Jiang, Ridong and Aw, AiTi},
  booktitle={Findings of the Association for Computational Linguistics: EMNLP 2020},
  pages={1063--1073},
  year={2020}
}

@article{cervone2020dialogue,
  title={Is this dialogue coherent? learning from dialogue acts and entities},
  author={Cervone, Alessandra and Riccardi, Giuseppe},
  journal={arXiv preprint arXiv:2006.10157},
  year={2020}
}

@article{cervone2018coherence,
  title={Coherence models for dialogue},
  author={Cervone, Alessandra and Stepanov, Evgeny and Riccardi, Giuseppe},
  journal={arXiv preprint arXiv:1806.08044},
  year={2018}
}

@article{zhang2021d,
  title={D-score: Holistic dialogue evaluation without reference},
  author={Zhang, Chen and Lee, Grandee and D’Haro, Luis Fernando and Li, Haizhou},
  journal={IEEE/ACM Transactions on Audio, Speech, and Language Processing},
  volume={29},
  pages={2502--2516},
  year={2021},
  publisher={IEEE}
}

@article{deriu2021survey,
  title={Survey on evaluation methods for dialogue systems},
  author={Deriu, Jan and Rodrigo, Alvaro and Otegi, Arantxa and Echegoyen, Guillermo and Rosset, Sophie and Agirre, Eneko and Cieliebak, Mark},
  journal={Artificial Intelligence Review},
  volume={54},
  pages={755--810},
  year={2021},
  publisher={Springer}
}

@article{santhanam2019towards,
  title={Towards best experiment design for evaluating dialogue system output},
  author={Santhanam, Sashank and Shaikh, Samira},
  journal={arXiv preprint arXiv:1909.10122},
  year={2019}
}

@article{chen2017survey,
  title={A survey on dialogue systems: Recent advances and new frontiers},
  author={Chen, Hongshen and Liu, Xiaorui and Yin, Dawei and Tang, Jiliang},
  journal={Acm Sigkdd Explorations Newsletter},
  volume={19},
  number={2},
  pages={25--35},
  year={2017},
  publisher={ACM New York, NY, USA}
}

@inproceedings{liu2017iterative,
  title={Iterative policy learning in end-to-end trainable task-oriented neural dialog models},
  author={Liu, Bing and Lane, Ian},
  booktitle={2017 IEEE Automatic Speech Recognition and Understanding Workshop (ASRU)},
  pages={482--489},
  year={2017},
  organization={IEEE}
}

@article{su2016line,
  title={On-line active reward learning for policy optimisation in spoken dialogue systems},
  author={Su, Pei-Hao and Gasic, Milica and Mrksic, Nikola and Rojas-Barahona, Lina and Ultes, Stefan and Vandyke, David and Wen, Tsung-Hsien and Young, Steve},
  journal={arXiv preprint arXiv:1605.07669},
  year={2016}
}

@article{zhao2021effective,
  title={Effective Sequence-to-Sequence Dialogue State Tracking},
  author={Zhao, Jeffrey and Mahdieh, Mahdis and Zhang, Ye and Cao, Yuan and Wu, Yonghui},
  journal={arXiv preprint arXiv:2108.13990},
  year={2021}
}

@article{kibble2004optimizing,
  title={Optimizing referential coherence in text generation},
  author={Kibble, Rodger and Power, Richard},
  journal={Computational Linguistics},
  volume={30},
  number={4},
  pages={401--416},
  year={2004},
  publisher={MIT Press}
}

@article{mckeown1997floating,
  title={Floating constraints in lexical choice},
  author={McKeown, Kathleen and Elhadad, Michael and Robin, Jacques},
  year={1997}
}

@article{popescu2009constraint,
  title={A constraint satisfaction approach to context-sensitive utterance generation in multi-party dialogue systems},
  author={Popescu, Vladimir and Caelen, Jean and Burileanu, Corneliu},
  journal={International Journal of Speech Technology},
  volume={12},
  pages={95--112},
  year={2009},
  publisher={Springer}
}

@article{moriceau2004constraint,
  title={A constraint-based model for preposition choice in natural language generation},
  author={Moriceau, V{\'e}ronique and Saint-Dizier, Patrick},
  journal={Constraint Solving and Language Processing},
  pages={124},
  year={2004}
}

@article{labruna2024you,
  title={Are you a Good Assistant? Assessing LLM Trustability in Task-oriented Dialogues},
  author={Labruna, Tiziano and Brenna, Sofia and Bonetta, Giovanni and Magnini, Bernardo},
  journal={Clic-It 2024},
  year={2024}
}

@article{hurst2024gpt,
  title={Gpt-4o system card},
  author={Hurst, Aaron and Lerer, Adam and Goucher, Adam P and Perelman, Adam and Ramesh, Aditya and Clark, Aidan and Ostrow, AJ and Welihinda, Akila and Hayes, Alan and Radford, Alec and others},
  journal={arXiv preprint arXiv:2410.21276},
  year={2024}
}

@article{dubey2024llama,
  title={The llama 3 herd of models},
  author={Dubey, Abhimanyu and Jauhri, Abhinav and Pandey, Abhinav and Kadian, Abhishek and Al-Dahle, Ahmad and Letman, Aiesha and Mathur, Akhil and Schelten, Alan and Yang, Amy and Fan, Angela and others},
  journal={arXiv preprint arXiv:2407.21783},
  year={2024}
}

@inproceedings{qin2023end,
    title = "End-to-end Task-oriented Dialogue: A Survey of Tasks, Methods, and Future Directions",
    author = "Qin, Libo  and
      Pan, Wenbo  and
      Chen, Qiguang  and
      Liao, Lizi  and
      Yu, Zhou  and
      Zhang, Yue  and
      Che, Wanxiang  and
      Li, Min",
    editor = "Bouamor, Houda  and
      Pino, Juan  and
      Bali, Kalika",
    booktitle = "Proceedings of the 2023 Conference on Empirical Methods in Natural Language Processing",
    month = dec,
    year = "2023",
    address = "Singapore",
    publisher = "Association for Computational Linguistics",
    url = "https://aclanthology.org/2023.emnlp-main.363/",
    doi = "10.18653/v1/2023.emnlp-main.363",
    pages = "5925--5941",
}

@inproceedings{cho-etal-2022-know,
    title = "Know Thy Strengths: Comprehensive Dialogue State Tracking Diagnostics",
    author = "Cho, Hyundong  and
      Sankar, Chinnadhurai  and
      Lin, Christopher  and
      Sadagopan, Kaushik Ram  and
      Shayandeh, Shahin  and
      Celikyilmaz, Asli  and
      May, Jonathan  and
      Beirami, Ahmad",
    editor = "Goldberg, Yoav  and
      Kozareva, Zornitsa  and
      Zhang, Yue",
    booktitle = "Findings of the Association for Computational Linguistics: EMNLP 2022",
    month = dec,
    year = "2022",
    address = "Abu Dhabi, United Arab Emirates",
    publisher = "Association for Computational Linguistics",
    url = "https://aclanthology.org/2022.findings-emnlp.391/",
    doi = "10.18653/v1/2022.findings-emnlp.391",
    pages = "5345--5359",
}

@article{BRAILSFORD1999557,
title = {Constraint satisfaction problems: Algorithms and applications},
journal = {European Journal of Operational Research},
volume = {119},
number = {3},
pages = {557-581},
year = {1999},
issn = {0377-2217},
doi = {https://doi.org/10.1016/S0377-2217(98)00364-6},
url = {https://www.sciencedirect.com/science/article/pii/S0377221798003646},
author = {Sally C. Brailsford and Chris N. Potts and Barbara M. Smith}
}

\end{document}